\pdfoutput=1

\documentclass[11pt]{article}

\usepackage{acl}

\usepackage{times}
\usepackage{latexsym}

\usepackage[T1]{fontenc}

\usepackage[utf8]{inputenc}

\usepackage{microtype}

\usepackage{caption}
\usepackage{subcaption}
\hypersetup{hidelinks}
\usepackage{breakurl}

\usepackage{balance}

\usepackage{amsmath}
\usepackage{booktabs}
\usepackage{amssymb}
\usepackage{graphicx}
\usepackage{color, colortbl}
\usepackage{xspace}

\definecolor{LightCyan}{rgb}{0.88,1,1}
\definecolor{LightRed}{rgb}{1.0, 0.91, 0.91}
\definecolor{LightGray}{rgb}{0.88,0.88,0.88}
\definecolor{VeryLightGray}{rgb}{0.93,0.93,0.93}

\usepackage{listings}
\newcommand{\id}[1]{{\sf\small #1}}
\newcommand{\myurl}[1]{{\small\url{#1}}}

\newcommand{\dsizeprecise}{133,817\xspace}
\newcommand{\dsize}{134k\xspace}
\newcommand{\dimages}{114k\xspace} 
\newcommand{\dimagesprecise}{113,971\xspace}

\newcommand{\dsites}{58k\xspace} 

\newcommand{\crawlsize}{104k\xspace}
\newcommand{\crawldata}{Mixed\xspace}

\newcommand{\officesize}{12k\xspace}
\newcommand{\officedata}{MS\xspace}

\newcommand{\zoomsize}{2.5k\xspace} 
\newcommand{\zoomdata}{Zoom\xspace}

\newcommand{\teamsize}{1.7k\xspace} 
\newcommand{\teamsdata}{Teams\xspace}

\newcommand{\androiddata}{Android\xspace}
\newcommand{\androidhowto}{AndroidHowTo\xspace}
\newcommand{\androidsize}{5k\xspace} 

\newcommand{\sys}{Lexi\xspace}
\newcommand{\dataset}{UICaption\xspace}
\newcommand{\roberta}{RoBERTa\xspace}
\newcommand{\vilbert}{ViLBERT\xspace}
\newcommand{\baseline}{\vilbert-UI\xspace}
\newcommand{\newbaseline}{\vilbert-12-1-UI\xspace}

\title{\sys: Self-Supervised Learning of the UI Language}

\author{Pratyay Banerjee$^1$\Thanks{The two authors contributed equally.}~\Thanks{Work done during an internship at Microsoft Research.} \hspace{1ex} Shweti Mahajan$^2$\footnotemark[1] \hspace{1ex} Kushal Arora$^3$\footnotemark[2] \hspace{1ex} Chitta Baral$^1$ \hspace{1ex} Oriana Riva$^2$
\AND
$^1$\textnormal{Arizona State University} \And \textnormal{$^2$Microsoft Research} \And \textnormal{$^3$McGill University}
\AND
}

\begin{document}
\maketitle

\begin{abstract}
Humans can learn to operate the user interface (UI) of an application by reading an instruction manual or how-to guide. Along with text, these resources include visual content such as UI screenshots and images of application icons referenced in the text. We explore how to leverage this data to learn generic visio-linguistic representations of UI screens and their components. These representations are useful in many real applications, such as accessibility, voice navigation, and task automation. Prior UI representation models rely on UI metadata (UI trees and accessibility labels), which is often missing, incompletely defined, or not accessible. We avoid such a dependency, and propose \emph{\sys}, a pre-trained vision and language model designed to handle the unique features of UI screens, including their text richness and context sensitivity. To train \sys we curate the \emph{\dataset} dataset consisting of \dimages UI images paired with descriptions of their functionality. We evaluate \sys on four tasks: UI action entailment, instruction-based UI image retrieval, grounding referring expressions, and UI entity recognition.
\end{abstract}

\begin{figure*}[t]
    \centering
    \begin{subfigure}[t]{0.34\textwidth}
    	\centering
    	\includegraphics[height=6cm]{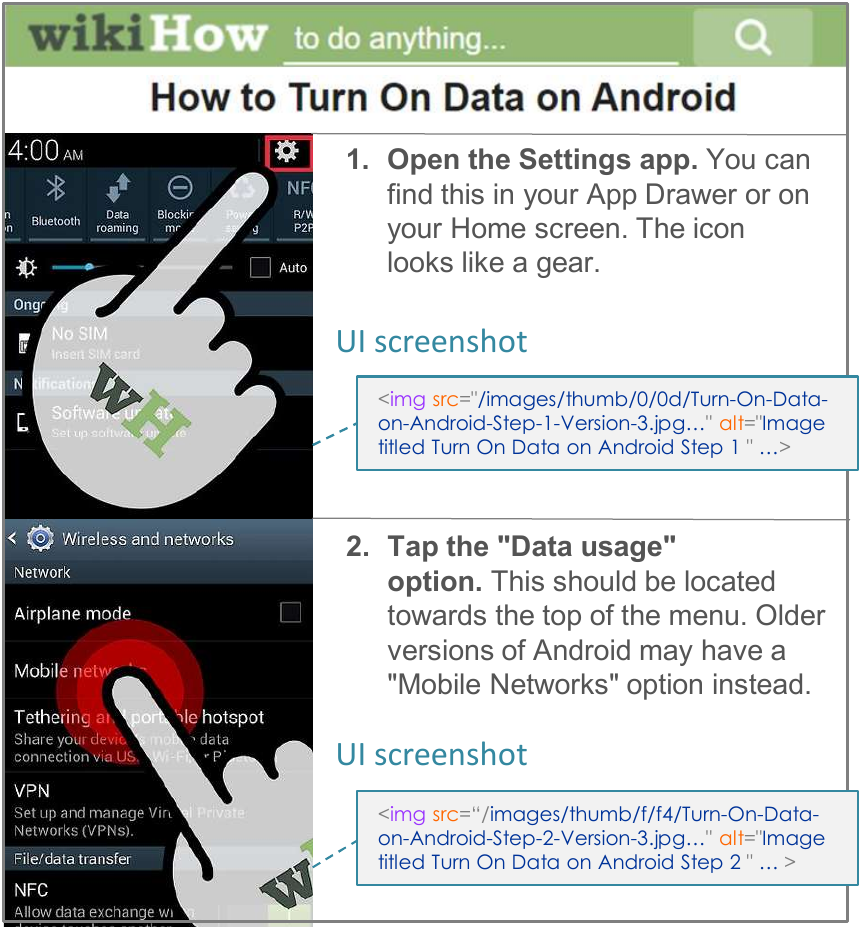}
    	\caption{WikiHow Android manual}
    	\label{fig:howto}
    \end{subfigure}
    \hfill
    \begin{subfigure}[t]{0.63\textwidth}
    	\centering
        \includegraphics[height=6cm]{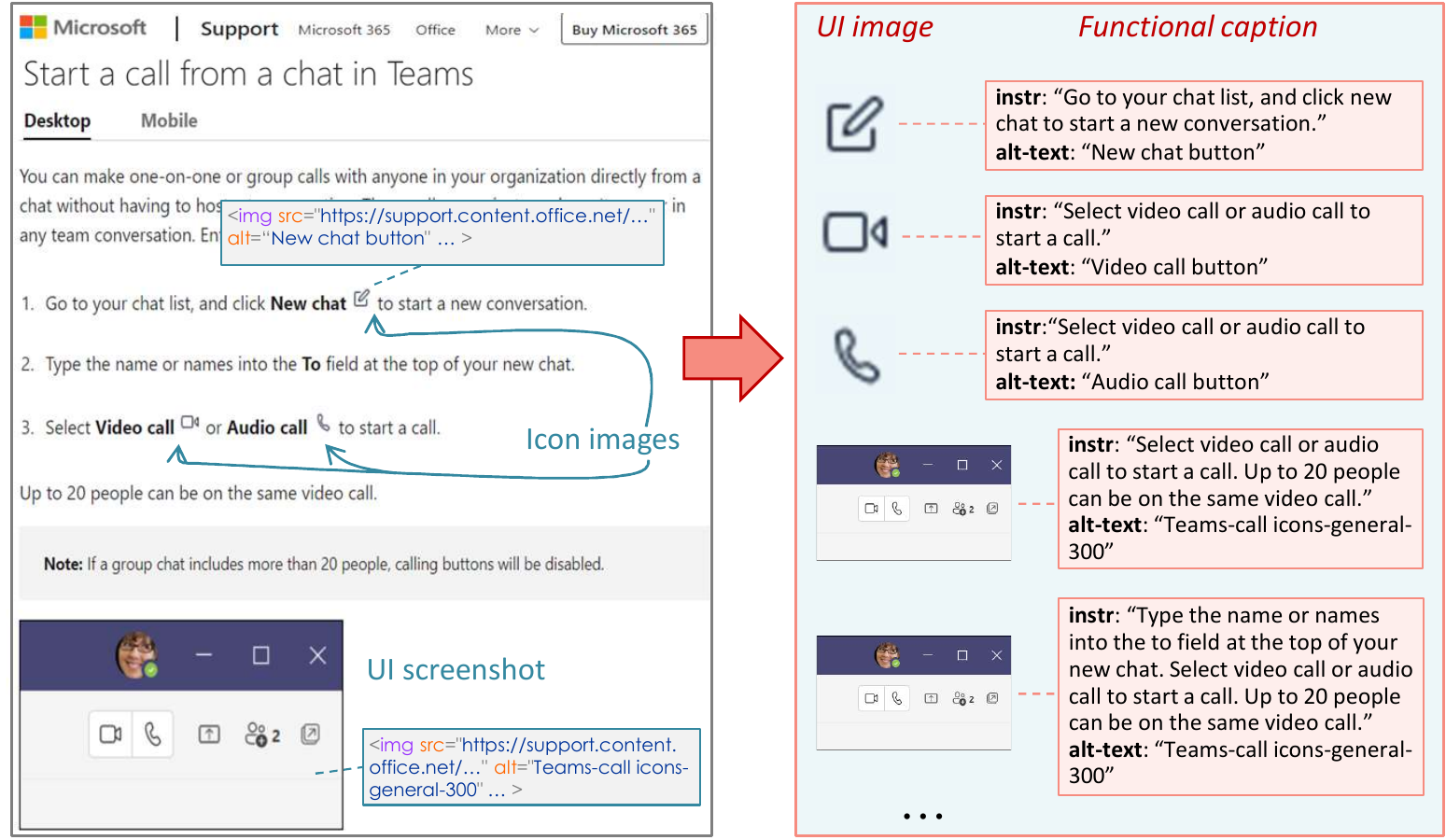}
        \caption{Microsoft Teams manual and generated image-caption pairs}
    	\label{fig:teams}
    \end{subfigure}
        \caption{Examples of how-to and support websites. In addition to text, they contain UI screenshots and icon images, which we use to generate image-caption pairs for training \sys.}
        \label{fig:manuals}
\end{figure*}

\section{Introduction}

Over the years, humans have learned to parse application user interfaces (UIs). With only a glance at a previously unseen UI they can understand its elements and identify those relevant to a desired task. To some extent, humans have learned the ``UI language''. The words of this language are UI elements, which can be visual (an icon), textual (a hyperlink), or both (a text button). If machines could parse and understand the UI language in the same way, they could map natural language commands to UI elements, facilitate access to UIs for visually-impaired users, and ultimately operate UIs on behalf of users. 

Towards this goal, we are interested in leveraging current vision and language (VL) representation models \cite[e.g.,][]{visual-bert19,lu2019vilbert,chen:uniter20} to achieve a generic visio-linguistic feature representation for UIs. However, some key constraints on data availability and data access need to be taken into account. Training VL models requires high-quality paired visio-linguistic datasets. Approaches to collect such datasets for UIs currently involve \emph{(i)} app crawling with human labelling~\cite{widget-caption,li-acl20}, and \emph{(ii)} use of unlabeled data consisting of UI screens and associated metadata~\cite{action-bert21,bai2021uibert}. While the former is hard to scale, in case of the latter, UI metadata, such as accessibility labels and structural representations of a UI screen (referred to as DOM tree in webpages and View Hierarchy in Android) is often missing~\cite{chen20:gui-accessibility}, partially defined, or not accessible for security reasons (on Android, only apps approved as Accessibility tools can access it~\cite{xda-ay11-block}).

To address these limitations,
we propose leveraging a new data source: 
instruction manuals, tutorials, and how-to guides that abound in the Internet (such as in technical support and how-to websites), exist for many different apps and platforms, and are easy to crawl. In addition to textual instructions, for clarity, they often include visual information. Figure~\ref{fig:manuals} shows two examples where step-by-step instructions from WikiHow include screenshots of the Android UI (left figure) and instructions from the Microsoft Teams manual include images of the referenced icons (middle figure). We posit that these visual and textual resources can be leveraged to learn visually-grounded textual representations of UI screens and their components.

Applying current VL models to such UI data raises some new challenges. VL models are trained using datasets like Conceptual Captions~\cite{sharma2018conceptual}, which contain images of the real world (i.e., photographs). In the physical world, objects of a certain class (cars, bikes, people) share certain appearance features, while the appearance of a UI element is less indicative of its meaning or functionality. UI elements may vary their appearance (in size, color, shape) but have the same functionality, or, vice versa, have different functionality but share similar appearance (e.g., a hand-shaped icon can mean ``Raise hand'' in a video conference app or ``Hand tool'' in Photoshop). In other words, the meaning of a UI element is highly \emph{context sensitive}. A second difference is that, compared to photographs, UI screens are \emph{rich in texts}; for some elements, such as text buttons or hyperlinks, text is essential. However, current VL models do not perform text recognition on the visual input. Finally, an 
out-of-the-box language model like BERT~\cite{devlin-etal-2019-bert}
is inadequate to represent manual instructions because \emph{(i)} highly-recurrent technical
terms (``click'', ``tap'', ``type'', etc.) may prevent it from
distinguishing instructions, and \emph{(ii)} manuals use terms (``menu'', ``bar'', ``button'', etc.) that have a very different meaning in the general domain.

To address these challenges we curate a new UI dataset (\emph{\dataset}) and propose \emph{\sys}, a pre-trained VL model for UI language understanding.

We crawl images of icons and UI screenshots from support and how-to websites, spanning multiple applications and platforms, and create a visio-linguistic dataset by pairing each image with one or multiple captions. To address the context sensitivity of UI elements, we synthesize \emph{functional captions} from instructions co-located with the image in the webpage. Rather than describing the appearance of an image, these captions describe the functionality of one or multiple UI elements, in the \emph{context} of other elements (see Figure~\ref{fig:teams}).

We build \sys by replacing the standard object detection methods (e.g., Faster R-CNN~\cite{ren2015faster}) used in current VL models with one specifically designed for UI screens. To leverage the text-richness of UIs, we augment an image's region features with textual features of the recognized texts. Finally, to accurately distinguish UI actions, objects, and input parameters in functional captions, we create a new pre-training task based on (noun-verb) POS tagging. We evaluate \sys on 4 downstream tasks: UI action entailment, instruction-based image retrieval, grounding referring expressions, and entity recognition.

In summary, we make the following contributions: \emph{(i)} a new data source for self-supervised learning of the UI language and a cross-platform dataset of \dsize pairs of images and functional captions, which we have released\footnote{\url{https://github.com/microsoft/UICaption}}; \emph{(ii)} \sys, a pre-trained model for UI language understanding which does \emph{not} depend on UI metadata; and \emph{(iii)} an evaluation of \sys, including baseline comparisons, out of domain performance, and ablation analysis.

\section{Related work}

\paragraph{VL models.} Multiple single- or two-stream transformer-based architectures have been proposed to learn a single feature space from visual and language inputs~\cite[e.g.,][]{tan2019lxmert,lu2019vilbert,visual-bert19,vl-bert20, chen:uniter20,imageBERT20}. Unlike pre-training of language models which can use unlimited natural language texts (e.g., Wikipedia), VL models use high-quality paired visio-linguistic datasets, such as Conceptual Captions~\cite{sharma2018conceptual}, SBU Captions~\cite{ordonez2011im2text}, and MS COCO~\cite{lin2014microsoft}. Our approach extends prior architectures with a vision encoder designed to leverage the text richness of UI images; we revisit VL models in a new problem space, UI language understanding, with a new set of data consisting of UI images and functional captions.

\paragraph{UI representation models.} \emph{Supervised} UI understanding methods depend on the collection of interaction traces, app crawling infrastructures, and human labeling. \citet{web-workflows18} and \citet{learning-navigate-web} use human demonstrations to train reinforcement learning agents that execute tasks specified in natural language by interacting directly with UIs. \citet{pasupat2018mapping} and \citet{li-acl20} curate datasets of natural language commands to learn how to ground language into low-level UI elements (buttons, icons, text inputs, etc.) and actions (click, type, select, etc.). All these systems learn at the level of UI elements which can be effective, but only in controlled environments (UI elements do not change over time~\cite{pmlr-v70-shi17a}) and for application-specific tasks~\cite{Branavan09}. For better generalization, \citet{etna-flin} map natural language commands to ``concept-level'' actions, but they require labeled natural language commands and a graph model of the execution environment. 

\emph{Unsupervised} methods include ActionBert~\cite{action-bert21} and UIBert~\cite{bai2021uibert}, consisting of transformer-based multimodal architectures trained on sequential or single Android UI screens, respectively. While these systems depend on the availability of UI metadata (UI trees and accessibility labels), \sys assumes only a screenshot is given, thus being widely applicable. By leveraging web data, which can be crawled without requiring app hosting infrastructures, \sys is trained to handle mobile and desktop UIs from multiple platforms.

\paragraph{Instruction manuals.} To the best of our knowledge we are the first using them to learn UI representation models. \citet{yagcioglu2018recipeqa} used them to study recipes in a multimodal setting, and \citet{bisk2020piqa} physical commonsense in a text-only setting. \citet{branavan-rl-game-2011} used them to inform the strategy of a game-playing agent.

\section{\dataset dataset} 
\label{sec:dataset}

We curate a new dataset of UI images (screenshots and icons) paired with functional captions.

We collect UI images, alt-text, and instructions accompanying a UI image from three sources: \emph{(i) Instruction manuals} extracted from support websites of 23 Microsoft (MS) apps (Teams, Outlook, Excel, Word, etc.) and Zoom. \emph{(ii) Android support websites}, whose URLs are obtained from the \androidhowto dataset~\cite{li-acl20}, and \emph{(iii) \dsites general tech websites} whose URLs are obtained by submitting search queries containing technical commands (extracted from MS and Android manuals) to Google image search and inspecting the top 60 results. To improve data quality, we perform data filtering, including a simple hash-based and text-based near-duplicate image removal, blurred image removal, and image relevance (e.g., excluding content-unrelated images such as Facebook and Twitter icons). 

For each collected image, we use the text surrounding it in the webpage to generate from 1 up to 5 \textbf{functional captions} (the number depends on the length and structure of the surrounding text). If available, we also append the alt-text. Rather than describing the appearance of an image, functional captions describe the functionality of one or multiple UI elements, in the \emph{context} of other elements. \emph{Textual} context may translate into \emph{spatial} context because the instructions to accomplish a task often involve co-located UI elements. For example, in Figure~\ref{fig:teams}, the ``Video call'' and ``Audio call'' icons are associated with the caption ``Select Video call or Audio call to start a call''; the instruction defines a logical relationship between the two functions, which corresponds to a spatial relationship, i.e., the two icons are next to each other in the UI.

We thus obtain the \textbf{\dataset dataset} consisting of \dimagesprecise unique UI images and \dsizeprecise image-caption pairs. This dataset significantly differs from those used in prior UI understanding models~\cite{action-bert21,bai2021uibert}. The data is extremely varied in terms of applications (Excel, GMail, Chrome, Photos, etc.), platforms (mobile \& desktop), number of UI elements (from single icons to complete desktops), text density (Word documents, photos, or Excel spreadsheets), pixel density, and quality (e.g., screenshots may not be pristine or may have been annotated, such as in Figure~\ref{fig:howto}). More details on the dataset and the crawling process can be found in Appendix~\ref{sec:appendix-dataset}.

\begin{figure*}[t]
    \centering
    \includegraphics[width=0.97\linewidth]{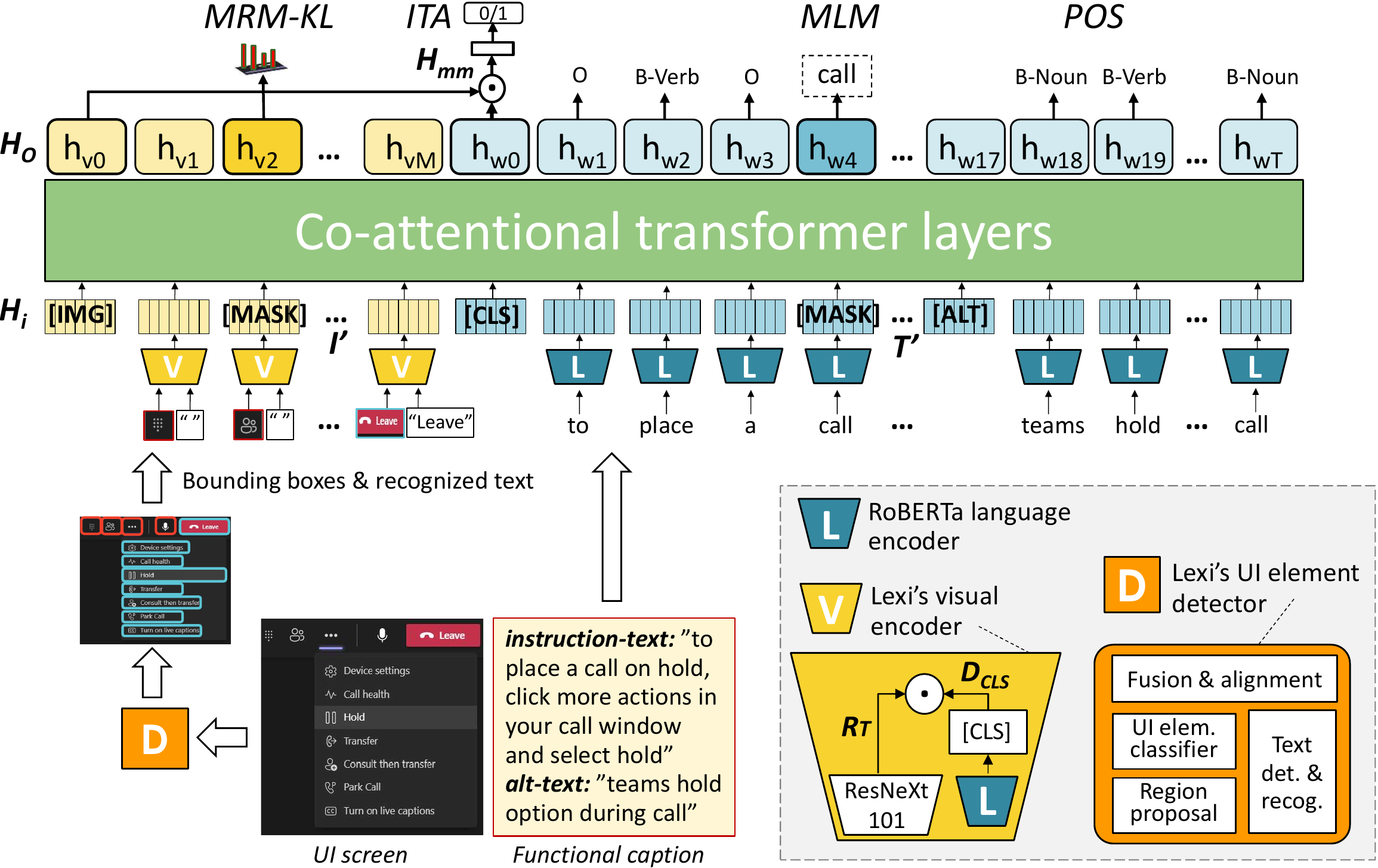}
    \caption{Our proposed VL transformer encoder, trained using 4 pre-training tasks: masked language modeling, masked region modeling with KL divergence, image-text alignment, and POS tagging.}
    \label{fig:arch}
\end{figure*}

\section{Method}
\label{sec:method}
\label{sec:scun}


\sys follows the two-stream transformer-based architecture of various VL models~\cite{lu2019vilbert, tan2019lxmert}. As illustrated in Figure~\ref{fig:arch}, it consists of an image encoder $V$, a text encoder $L$, and co-attentional transformer layers for cross-modal representation learning. Given a text input $T$ represented as a set of word tokens $w_1, ..., w_T$ and an image $I$ represented as a set of region features $v_1,..., v_M$ augmented with their corresponding detected texts, \sys outputs final representations $h_{v0}, ..., h_{vM}$ for the visual stream and $h_{w0},..., h_{wT}$ for the textual stream. The most distinct aspect of \sys is the early-infusion of detected text vector representations with corresponding visual features, which is motivated by the frequent presence of text in UI images. Moreover, the vocabulary of functional captions including technical action verbs and specific named entities motivates an additional self-supervision task for pre-training.

\paragraph{UI image representation.} An input image $I$ is first resized to a fixed size (224$\times$224), and encoded by extracting bounding boxes of its regions of interest and corresponding visual features $v$. To this end, VL architectures~\cite{lu2019vilbert} rely on pre-trained object detectors (e.g., Faster R-CNN~\cite{ren2015faster} pre-trained on Visual Genome~\cite{krishna2017visual}). Pre-training (or fine-tuning) object detectors for UI images requires large-scale datasets of annotated UI screens, which are generally not available.\footnote{Only the Android Rico dataset exists~\cite{deka2017rico}.} Inspired by~\citet{chen2020object}'s work, we implement a UI element detection framework, the \emph{\sys detector}, which can handle the cross-platform and cross-app nature of the UI images in our dataset, without requiring annotated UI screens. The design is optimized to leverage the regular layouts and text richness of UI screens. In short, it works as follows. It detects bounding boxes of UI elements using standard computer vision techniques (Canny edge detection~\cite{canny86}, contour detection~\cite{SUZUKI1985-contour-map}, non-maximum suppression) and classifies them using a ResNeXt-101 model fine-tuned on a 9-class dataset of 263k mobile/desktop UI elements.\footnote{We derive the dataset from icon collections~\cite{google-material-icons} and mobile app datasets~\cite{deka2017rico}, and through data synthesis based on UI markup files of Windows apps.} In parallel, it uses OCR (PP-OCR~\cite{du2020pp}) to detect text elements and recognize their text. Then, it combines the two types of prediction by merging boxes based on alignment and distance, to output a set of classified bounding boxes with recognized text, if any. (Appendix~\ref{sec:ui-elem-detector} provides further details on the implementation.)

For each image $I$, \sys's vision encoder (Figure~\ref{fig:arch} bottom-right) computes region features using our fine-tuned ResNeXt-101 model. It keeps a maximum of 64 high-scoring boxes, and for each selected region $i$, it computes $v_i$ (of size 2048) as the mean-pooled convolutional feature from $i$. In state-of-the-art VL models, text appearing in an input image is not directly extracted. To leverage the text richness of UI screens, each $v_i$ is augmented with textual features of the recognized text (computed as dot-product). Textual features help align with functional captions (as demonstrated by the performance of our baseline \baseline-dt, \S\ref{sec:downstream-tasks}). Finally, we encode the spatial location consisting of region position and fraction of the image area covered (not shown in Figure~\ref{fig:arch}). This is projected to match the dimension of the visual/textual features and they are summed. We mark the beginning of the image stream with a special \textsc{[IMG]} token representing the entire image.

\paragraph{UI text representation.}
Text is encoded as in BERT~\cite{devlin-etal-2019-bert}, as a sequence of word tokens pre-pended with the \textsc{[CLS]} token. We concatenate the image's associated instructions and alt-text using a new special token \textsc{[ALT]}. For a given word token, the input representation is a sum of a token, segment, and position embedding. The text encoder is initialized using \roberta~\cite{liu2019roberta}. We tokenize the text input using a byte-level BPE~\cite{sennrich-etal-2016-neural} tokenizer.



%

Formally, given the text $T = w_1, ..., w_T$, the image $I = v_1, ..., v_M$ and the corresponding detected texts $D_{v} = d_{v1}, ..., d_{vM}$, the image encoder $V$ and the text encoder $L$, \sys is represented as follows:
\vspace{-1.2ex}
\begin{equation*}
\begin{split}
& I' = V(v_i) = D_\textsc{{CLS}}(v_i) \cdot \textsc{Rt}(v_i), ~~T' = L(w_i)\\
& H_i = \id{[IMG]} : I' : T' \\
& H_o = \textsc{Co-Att}(H_i) \\
& H_{mm} = H_{v0} \cdot H_{w0}
\end{split}
\end{equation*}

\vspace{-1.2ex}
where, $D_{CLS}(v_i) = L(d_{vi})$ is the [\id{CLS}] token representation of the detected text in image I, $\textsc{Rt}$ is the fine-tuned ResNeXt-101 visual feature extractor whose weights are frozen, $T'$ is the encoded text input,
$H_i$ is the final input vector using concatenation operator $:$, and $H_o$ is the final output vector after the co-attentional transformer layers \textsc{Co-Att}. \textsc{Co-Att} takes as input the vector representations of the textual and visual streams projected to equal dimensions using linear layers. \textsc{Co-Att} is borrowed from \vilbert, in which the keys and values of an attention-block from each modality are passed as input to the other modality's multi-headed attention. This leads to attention-pooled features for each modality conditioned on the other, i.e, language-conditioned visual features and visual-conditioned language features. 
$H_{mm}$ is the final multi-modal hidden representation after a dot-product between the output representation of [\id{IMG}] of the visual input $I$ and [\id{CLS}] of the textual input $T$. The hidden representations from $H_o$ are further fed through respective task-specific feed-forward layers.


\paragraph{Pre-training tasks.} We train \sys using four pre-training tasks, three borrowed from prior work~\cite{devlin-etal-2019-bert,lu2019vilbert} and one new.

\emph{Masked Language Modeling (MLM).} We borrow this task from BERT. We randomly mask out input words with 15\% probability, and replace them with the \id{[MASK]} token in the text stream $T'$. We task the model with predicting them by leveraging the surrounding words and the visual input. We use standard cross-entropy loss for this task.




\emph{Masked Region Modeling with KL Divergence (MRM-KL).} We mask an image's regions with 50\% probability (and only if the image contains at least 4 regions). To mask image regions, we replace their features with zeros -- we mask 10\% of the features and retain 90\% of them. If a masked region contains text, the detected text is automatically masked out too. The model is tasked with reconstructing the masked regions given the remaining regions and the input words. However, the model is unlikely to be able to reconstruct exact image features. Hence, as in \vilbert, the model predicts a distribution over semantic classes for the corresponding image region, and minimizes the KL divergence between the original class probabilities of the unmasked region and the predicted class probabilities. We use cross-entropy loss also for this task.


\emph{Image-Text Alignment (ITA).} 
This task is useful for learning representations for downstream cross-modal tasks. It is a binary classification task: the cross-modal representation $H_{mm}$ is fed into a feed-forward layer to predict a score between 0 and 1. In training, we sample a positive or negative pair at each step. The negative pair is created by pairing the true image with a random caption that is at $[0.5-0.9]$ distance away in cosine-similarity with the true caption -- due to the prevalence of common technical terms, functional captions tend to be semantically very similar. We apply binary cross-entropy loss to this task. 





\emph{POS Tagging (POS).} A challenge with our text embeddings is that our text inputs have many technical terms, which makes it hard to distinguish them. Some terms (``home'',``bar'', etc.) also have a different meaning in general domain. Moreover, in current VL models the attention in the cross-modality layers focuses on nouns and pronouns because they are the most informative words in current datasets/tasks~\cite{tan2019lxmert}, but in our domain verbs play a key role. To help the model learn the ``foreign'' UI language and attend its syntax correctly, we introduce weakly-supervised part-of-speech (POS) tagging as a new pre-training task, where the model must predict noun and verb spans present in the text stream. 
Prior work uses POS features as input to a neural network~\cite{kiperwasser2016simple}, however, we aim to encode this information in the learned representations during pre-training. We use spaCy's POS tagger's predictions as weak-supervision labels for learning the task, and adopt the \id{BIO} scheme. Each token representation is classified as the beginning (\id{B}) or the inside (\id{I}) of a noun or verb entity, or as other (\id{O}), i.e., a non-entity token. We use cross-entropy loss.

\vspace{1ex}

\noindent In summary, the total pre-training loss $\mathcal{L}$ with scalar coefficients $\alpha,\beta,\gamma,\delta \in (0, 1]$ is given by:
\vspace{-1.5ex}
\begin{equation*}
\resizebox{\hsize}{!}{
    $\mathcal{L}$ = $\alpha\cdot\mathcal{L}_{MLM} + \beta\cdot\mathcal{L}_{MRM-KL} + \gamma\cdot\mathcal{L}_{ITA} + \delta\cdot\mathcal{L}_{POS}$
}
\end{equation*}

\vspace{-1.2ex}
where $\mathcal{L}_{x}$ is the loss for the pre-training task $x$.

\section{Evaluation on downstream tasks}
\label{sec:downstream-tasks}

We evaluate \sys through 4 downstream tasks. In this section, we describe our experimental setup, introduce datasets and baselines, describe each task's settings and results, and provide an ablation study.

\subsection{Experimental setup}

To pre-train \sys we apply the four tasks described in \S\ref{sec:method} to the \dataset dataset, excluding the Zoom and Teams portions which are used to conduct an out-of-domain (OOD) analysis. We create training (103k UI images, 116k image-text pairs ), validation (5k images, 5.6k image-text pairs), and test (6k images, 7k image-text pairs) splits by ensuring no overlaps between UI images and functional captions. In fine-tuning we use a subset of the pre-training dataset and two Android datasets, UIBert~\cite{bai2021uibert} and \androidhowto~\cite{li-acl20}.

We test two variants of \sys with 6 (\textbf{\sys-6}) or 10 (\textbf{\sys-10}) transformer layers in the visual stream. We train on 8 Nvidia V100 GPUs for 100 epochs with a batch size of 128 and a learning rate of 4e-5. We use the Adam optimizer~\cite{kingma2014adam}. For more details see Appendix~\ref{sec:training-details}.



\subsection{Baseline models}


We compare \sys against three baselines. We modify the \vilbert architecture to encode the input image using the \sys detector (instead of Faster R-CNN which is not trained on UI images) and the input text using \roberta. We test two variants of this baseline depending on whether the text detected in the image is encoded (\textbf{\baseline-dt}) or not (\textbf{\baseline}). The third baseline, \textbf{\newbaseline}, is a multi-task \vilbert model trained on 12 different datasets (4.5M instances) and 6 tasks~\cite{vilbert12-1}; it uses the \sys detector for image encoding and BERT for text. We fine tune all baselines on our fine-tuning datasets. In the entity recognition task (\S\ref{sec:ui-entity-task}), we introduce also a text-only baseline, \textbf{\roberta-UI}, the \roberta model fine-tuned on our dataset. We do not consider single-modality baselines in the other tasks as they cannot produce an output unless both types of input are given. The ActionBert~\cite{action-bert21} and UIBert~\cite{bai2021uibert} models require UI metadata (Android View Hierarchy and accessibility tags) as input, hence they are not directly comparable with \sys (and their source code is not released). However, wherever appropriate, we refer to their performance as upper bounds.

We test in zero-shot (ZS), fine-tuned (FT), and OOD settings. In ZS we directly apply the pre-trained model (i.e., the model is new to the task) to UI data not used in pre-training. In OOD, we test on data from the Zoom and Teams manuals. Both these apps are excluded from pre-training and fine-tuning, and, in general, when crawling \dataset no video conferencing app was specified in the search queries, making this app category completely unknown to the pre-trained model.


\begin{figure}[t]
    \centering
    \begin{subfigure}[t]{0.21\textwidth}
    	\centering
    	\includegraphics[height=4.2cm]{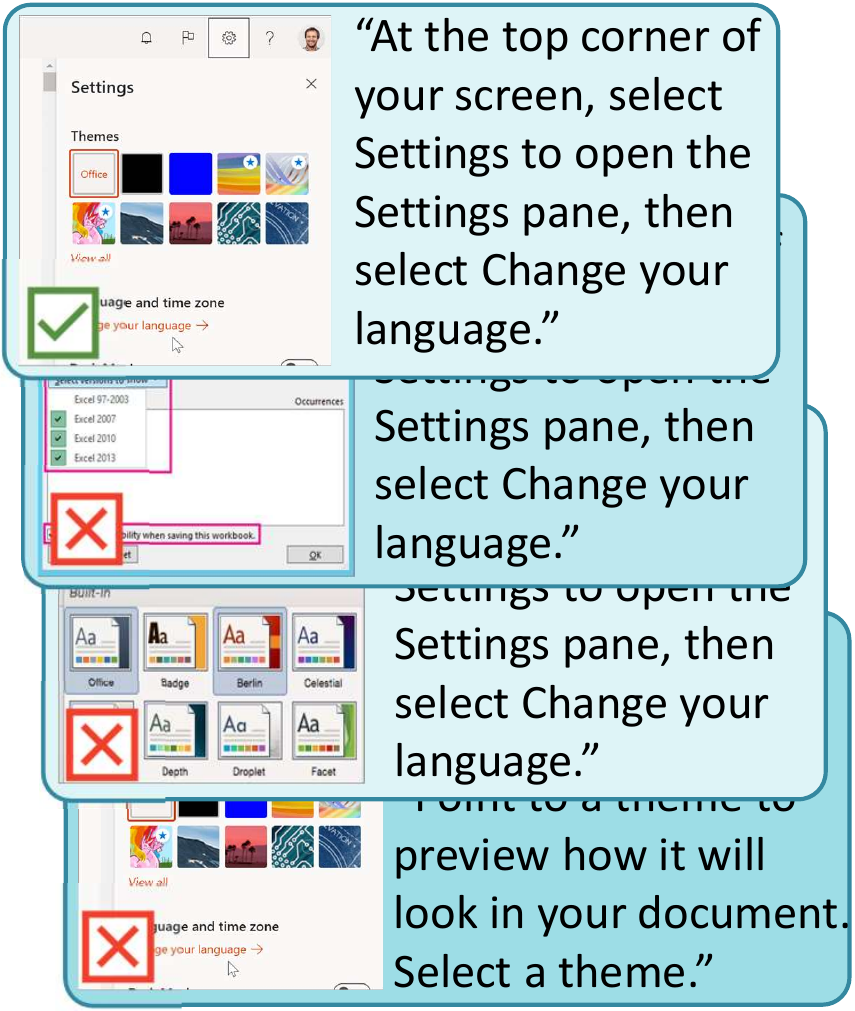}
    	\caption{Action entailment task.}
    	\label{fig:ae_task}
    \end{subfigure}
    \hspace{2ex}
    \begin{subfigure}[t]{0.22\textwidth}
    	\centering
        \includegraphics[height=4.2cm]{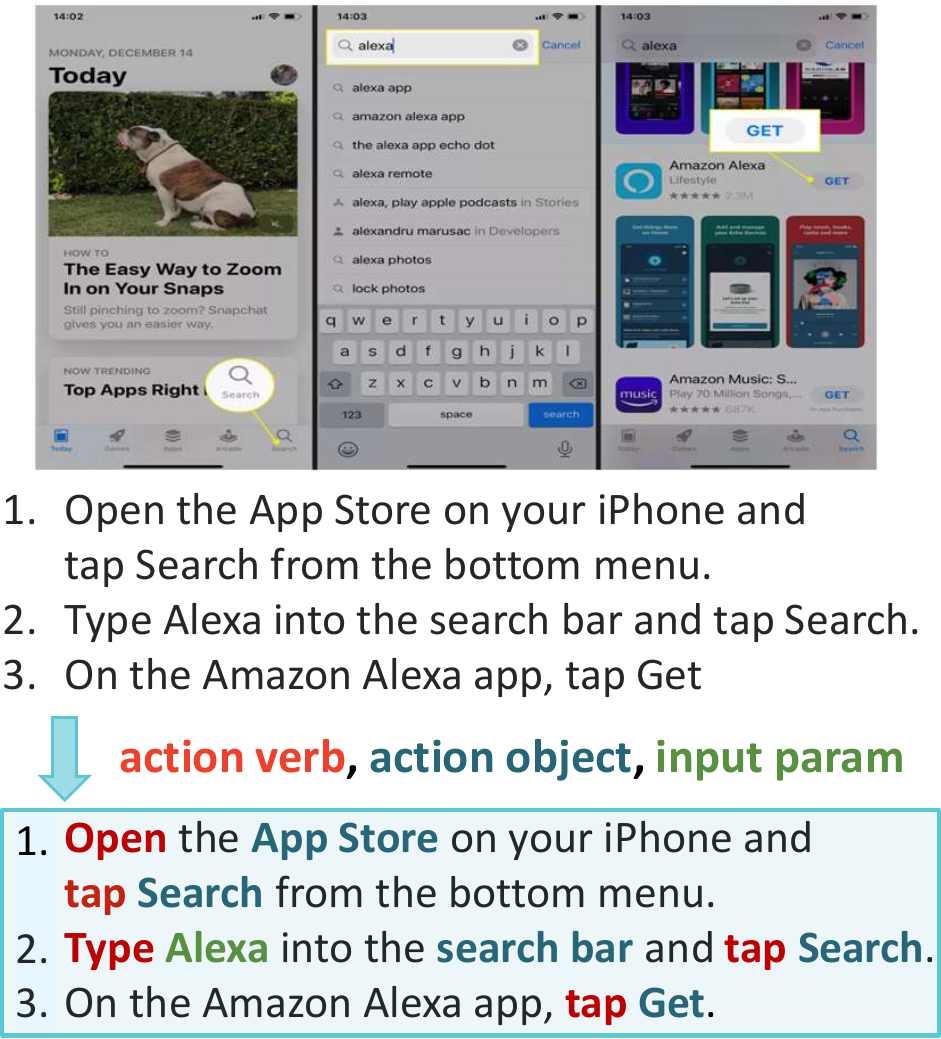}
        \caption{Entity recognition task.}
    	\label{fig:er_task}
    \end{subfigure}
        \caption{Examples of two \sys's downstream tasks.
        }
        \label{fig:2tasks}
\end{figure}

\begin{table*}[t]
\centering
\scalebox{0.85}{
\begin{tabular}{@{}lcccccccccccc@{}}
\toprule
& \multicolumn{3}{c}{UI action entailment} & \multicolumn{9}{c}{Instruction-based UI image retrieval} \\
\cmidrule(lr){2-4} \cmidrule(lr){5-13}
Model & \dataset & T (OOD) & Z (OOD) & \multicolumn{3}{c}{\dataset} & \multicolumn{3}{c}{T (OOD)} & \multicolumn{3}{c}{Z (OOD)}  \\ 
 & Acc \% & Acc \% & Acc \% &  R1 & R5 & R10 & R1 & R5 & R10 & R1 & R5 & R10  \\
\midrule
\baseline       & 65.0 & 31.6 & 30.7 & 12.9 & 42.7 & 59.3 & 4.1 & 17.0 & 32.0 & 3.3 & 16.8 & 29.5  \\
\baseline-dt    & 68.0 & 31.8 & 34.2 & 14.4 & 47.1 & 64.2 & 5.1 & 17.4 & 33.8 & 3.5 & 17.6 & 31.1  \\
\newbaseline    & 66.4 & 33.8 & 27.6 & 19.0 & 52.4 & 72.7 & 3.4 & 16.2 & 32.0 & 2.9 & 12.5 & 26.8 \\
Lexi-6 (ZS)     & 52.2 & - & - & 12.4 & 21.7 & 29.3 & - & - & - & - & - & -    \\ 
Lexi-10 (ZS)    & 57.4  & - & - & 15.0 & 22.2 & 29.0 & - & - & - & - & - & -   \\
Lexi-6 (FT)     & 83.3 & \textbf{40.1} & \textbf{39.6} & 34.7 & 78.4 & 89.1 & \textbf{6.5} & 25.1 & 44.3 & 4.7 & \textbf{22.7} & \textbf{39.9}  \\ 
Lexi-10 (FT)    & \textbf{84.1} & 39.3 & 36.1 & \textbf{36.7} & \textbf{80.0} & \textbf{91.8} & 4.7 & \textbf{27.5} & \textbf{45.8} & \textbf{5.3} & 21.5 & 38.2  \\
\bottomrule
\end{tabular}
}
\caption{Performance of \sys on the tasks of UI action entailment and instruction-based UI image retrieval. Models are used zero-shot (ZS) or fine-tuned (FT) on a 77k subset of \dataset. OOD tests are performed on Teams (T) and Zoom (Z).} 
\label{tab:ae-ir}
\end{table*}

\subsection{UI action entailment}
\label{sec:act-ent}

This task is inspired by the visual entailment task~\cite{Xie2019VisualEA}. Given a natural language instruction (e.g., ``Select video call to start a call'') and an image (screenshot or icon), the goal is to predict whether the described action can be performed on the image. This task has a practical use in screen readers to locate UI components by their name/functionality in the absence of accessibility labels and alt-texts, as it is often the case. While this task is similar to the ITA task used in pre-training, it is a useful diagnostic task to verify \sys has been pre-trained effectively.


In fine-tuning we use a 4-way multiple-choice setting (Figure~\ref{fig:ae_task}). Given an instruction and UI image (the true pair), we generate 3 negative pairs by sampling a hard negative from the 1,000 most similar images to the target image\footnote{obtained based on the Euclidean distance of their embeddings, and by discarding images whose captions are contained in the true caption or have a similarity below 0.5 or above 0.9.}, by substituting the true instruction with a random one and the true image with a random one. For each option we compute the alignment score (as in pre-training) and apply a softmax. To make the task more challenging but also more realistic, the image's alt-text is not provided in the input. For training we use a subset of the \dataset dataset used in pre-training (train:77k, val:3.7k, test: 6k\footnote{The test set was not seen in pre-training.}).

As Table~\ref{tab:ae-ir} shows, \sys (FT) largely outperforms the strongest baseline by 24\% demonstrating the efficacy of our pre-training tasks. In ZS settings, \sys achieves 57.4\% accuracy, demonstrating that the model has developed some ability to ground text without requiring any fine-tuning and without relying on the image's alt-text. We test OOD using 1.7k pairs from Teams (T) and 2.5k pairs from Zoom (Z), which were not included in fine-tuning and pre-training. \sys achieves 39.6--40.1\% accuracy (a 16--19\% gain over baselines). \sys-6 outperforms \sys-10 in the OOD settings, suggesting the model may be able to generalize better with a smaller network depth. In Appendix~\ref{sec:appendix-visual-data}, we provide a qualitative analysis of the OOD predictions.


\subsection{Instruction-based UI image retrieval}

This task is inspired by the image retrieval task used to test VL models. The goal is to identify a UI image from a pool of 50 given a description of its associated instruction(s). We fine-tune the models using a 4-way multiple-choice setting as for UI action entailment (but with alt-text included) on the same 77k subset of \dataset. In-domain and out-of-domain test sets are as in the UI action entailment task.

In Table~\ref{tab:ae-ir} we report recall@k with k=1,5,10. Our pre-training tasks are effective with significant gains in the FT settings in in-domain tests ($1.9\times$ higher R@1 than the \newbaseline baseline). While the OOD performance is lower, compared to the strongest baseline R@1 is still 1.4--1.8 percentage points higher. The best ZS performance, achieved by \sys-10, is inferior to that of the fine-tuned baseline models, indicating it is challenging for the pre-trained model to generalize to this new task. As observed in the previous task, increasing the number of visual layers has a positive effect on in in-domain tests, but the OOD performance of \sys-6 is better than that of \sys-10 in some cases. 

\begin{figure}[t]
    \centering
    \includegraphics[width=1.01\linewidth]{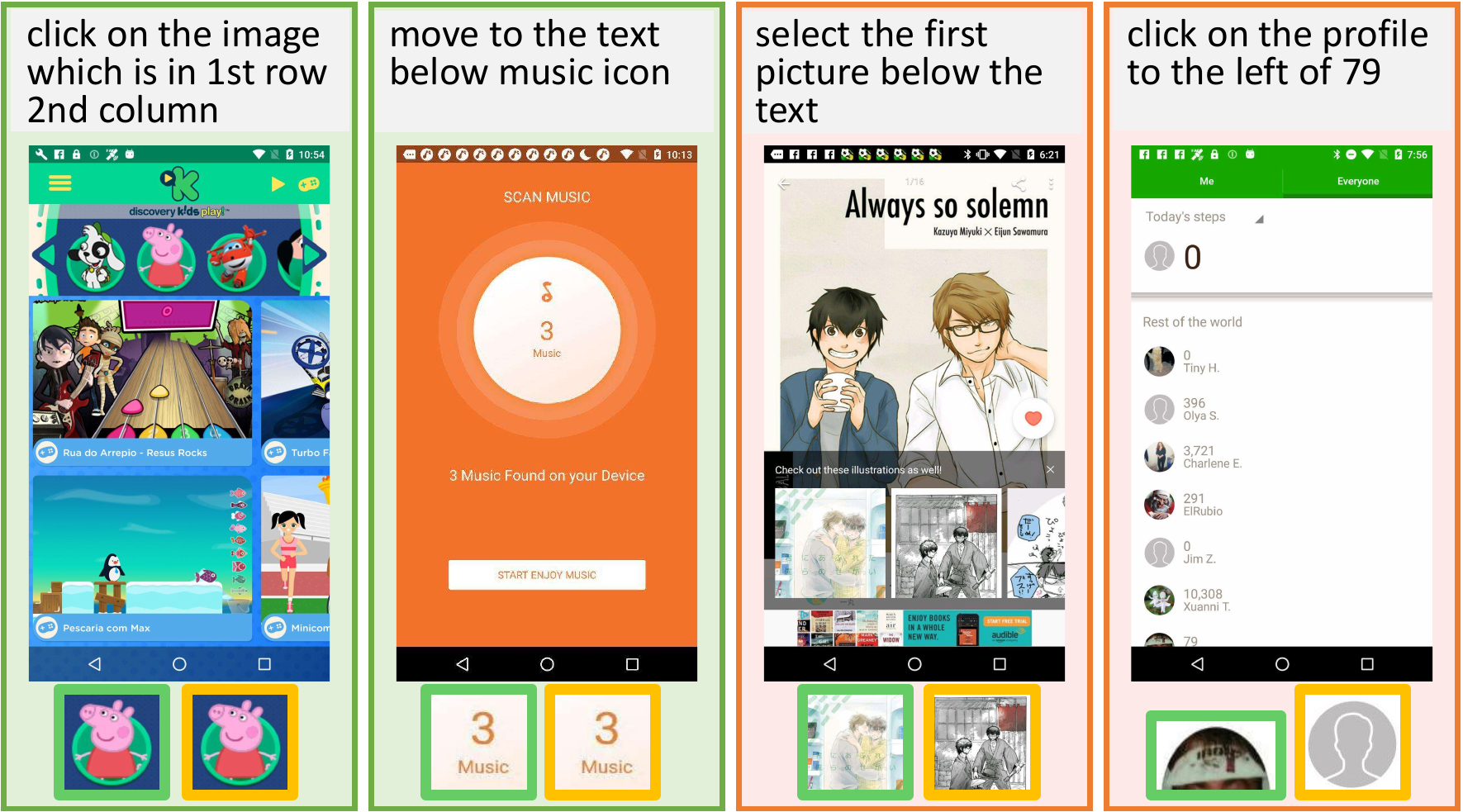}
    \caption{Examples of correct (first two) and wrong (last two) predictions for the grounding referring expressions task. The referring expression is above the UI screen, and ground truth UI elements (with a green border) and \sys's predictions (with an orange border) are below.}
    \label{fig:error-analysis}
\end{figure}

\subsection{Grounding referring expressions}

We borrow this task from ActionBert and UIBert. Given a referring expression in natural language and a UI screen, the goal is to select the UI element referenced by the expression from a set of elements detected in the screen. This task is relevant to voice-guided app navigation agents, where users may issue commands such as ``click home at the top of the page'' or ``play the third song in the list''. 

Given the referring expression, we re-rank a set of image region proposals. We pass the final representation $h_vi$ for each region $v_i$ into a learned linear layer to predict a matching score for fine-tuning. As we are given pre-computed image regions, the highest-scoring region is used as the prediction. We train using cross-entropy loss on the UIBert dataset, using the official splits (train:15.6k, val:448, test:544). On average, for each expression the model must choose from 20 candidate UI elements.


\sys outperforms the best baseline by 6.4\% (Table~\ref{tab:uiner}). If accessibility labels had to be available, the performance would likely increase, as demonstrated by UIBert that, with UI metadata, achieves 90.8\% accuracy. Figure~\ref{fig:error-analysis} provides a qualitative analysis of some predictions (2 correct and 2 wrong). The task is challenging, and the model makes reasonable errors. For instance, for the third referring expression, \sys correctly returns a picture, but the second rather than the first one under the text. In the last example, \sys correctly returns a profile icon but not the one next to the text ``79'', which, slightly cropped at the very bottom of the screen, may be hard to locate also for a human .

\begin{table}[t]
\centering
\resizebox{0.9\linewidth}{!}{
\begin{tabular}{@{}lcccc@{}}
\toprule
& Ref Exp & \multicolumn{3}{c}{UI Entity Rec} \\
\cmidrule(lr){2-2} \cmidrule(lr){3-5}
Model & UIBert & \multicolumn{3}{c}{MS+Android}  \\
& Acc \% & Prec & Recall & F1 \\
\midrule
RoBERTa-UI & - & 75.2 & 80.2 & 77.2 \\
\baseline & 51.7 & 75.7 & 81.0 & 77.8 \\
\baseline-dt & 58.5 & 76.5 & 80.1 & 77.8 \\
\newbaseline & 62.9 & 79.0 & 80.4 & 79.2 \\
Lexi-6 & \textbf{66.9} & 79.0 & \textbf{82.1} & \textbf{80.2} \\
Lexi-10 & 65.8 & \textbf{79.1} & 81.7 & 80.0 \\ \bottomrule
\end{tabular}
}
\caption{Performance of \sys on the grounding referring expressions and UI entity recognition tasks. Models are fine-tuned on UIBert and \officedata+Android, respectively.}
\label{tab:uiner}
\end{table}

\subsection{UI entity recognition}
\label{sec:ui-entity-task}

This task is relevant to task completion systems, to translate natural language commands into executable scripts, and was inspired by the AndroidHowTo dataset~\cite{li-acl20}.\footnote{In the tests we do not consider their model as a baseline because, although related, its goal and metrics are different (entity extraction vs. sequence generation, see Appendix~\ref{sec:tuple-extr} for more details).} Given a UI image and an instruction, the goal is to extract spans of \emph{UI action verbs}, \emph{UI action objects}, and \emph{input parameters} (see example in Figure~\ref{fig:er_task}). UI action verbs describe operations (click, type, etc.) that a user can take on a UI object (a button or a menu); some UI objects (a text field) require input parameters (e.g., ``Enter \emph{a} in b''). A challenge here is that not all instances of a word are of the same entity type (in Figure~\ref{fig:er_task}, ``Alexa'' is an input, but ``Alexa app'' is not; ``Search'' is an object, but may be a verb elsewhere). We model the task as a named-entity recognition task with \id{Begin (B)}, \id{Intermediate (I)}, and \id{Other (O)} class annotations (\id{BIO} scheme). We use a cross-entropy loss function.



We build a dataset for this task as follows. We ask 4 annotators to label the span of words that describe these entity types in instructions extracted from the manuals of 14 \officedata apps (more details on the process in Appendix~\ref{sec:appendix-annotation}). We combine our annotated instructions with similarly-annotated \androidhowto commands adapted to our task (Appendix~\ref{sec:tuple-extr}) to obtain 43.4k unique image-instruction pairs, for a total of 76.7k action verbs, 79.3k objects, and 1.5k inputs labelled. We refer to this dataset as MS+Android. For training and testing we use an 80-10-10 scheme.


As Table~\ref{tab:uiner} shows, \sys outperforms the \roberta baseline by 3.9\% and even the very large \newbaseline model by 1.3\% (based on F1 scores), thus demonstrating the effectiveness of pre-training with POS tagging. Learning with a multi-modal setup improves performance, suggesting the visual input provides further necessary context.

\subsection{Ablation analysis}
\label{sec:ablation}

We perform an ablation analysis where \sys-6 does not include one of our four pre-training tasks or does not use the text detected in the input image. All \sys variants are pre-trained as the complete \sys model. In the analysis we use the UI action entailment and UI entity recognition tasks, for which models are fine-tuned on a subset of \dataset (77k) and MS+Android, respectively. As Table~\ref{tab:ablations} shows, among the 4 pre-training tasks, ITA is the most relevant to UI action entailment (its absence causes a 17.8\% drop in \sys-6's accuracy) while MLM and POS are the least relevant. On the other hand, for UI entity recognition, \sys-6 slightly outperforms \sys-10, and we observe how POS and MRM-KL tasks are important to achieve high performance (their absence causes a 2.6\% drop in F1 scores). The inclusion of the detected text is most beneficial to UI action entailment. 

\begin{table}[t]
\centering
\scalebox{0.82}{
\begin{tabular}{@{}lcc@{}}
\toprule
Model & UI action ent  & UI entity rec \\ 
& Acc \% & F1 \\
\midrule
\baseline & 65.0 & 77.8 \\
\baseline-dt & 68.0 & 77.8 \\
\newbaseline & 66.4 & 79.2 \\
\sys-10 & \textbf{84.1} & 80.0\\  
\sys-6 & 83.3 & \textbf{80.2} \\
-- MLM & 83.2 & 78.6 \\
-- MRM-KL & 81.7 & 78.2 \\
-- ITA & 70.7 & 79.1 \\
-- POS & 82.8 & 78.2 \\ 
-- DetText & 75.8 & 79.2 \\
\bottomrule
\end{tabular}
}
\caption{Ablation study of \sys with respect to the UI action entailment and UI entity recognition tasks.}
\label{tab:ablations}
\end{table}

\section{Limitations}
In this section, we describe some limitations of our method, dataset, and data collection approach. 

Our approach depends on pre-training using publicly available web data, which is crawled at scale and for a diverse set of applications. Web data is highly skewed towards popular applications; less popular application might have less or lower-quality documentation and images online. In the future, we will focus on collecting UI data from other sources, such as instructional videos, online tutorials, and bug reports. We also observe that for certain application categories, such as shopping, transportation, weather, or news, instruction manuals are rare or totally missing. 

Another limitation that arises from curating pre-training data from the web is how to ensure zero overlap between pre-training and downstream evaluation tasks. A URL-based filter of the crawled websites does not eliminate overlaps as aggregator websites like wikiHow and LifeWire may contain manuals of different apps and platforms. Partitioning data based on application names is also difficult as the name of the application is not explicitly encoded in the webpage source.

Our data and method have been evaluated for English instructions and UIs that contain text in English. Future work should expand to different languages. Currently, the method is trained and evaluated on instructions whose length is at most 512 tokens. 

Pre-training the model even on a modest dataset like \dataset already requires significant GPU resources. The obtained model is not optimized to run on edge devices like mobile phones, thus possibly raising privacy concerns. For some scenarios that motivate this work (e.g., accessibility support and voice-based navigation in mobile apps) it would be highly desirable to execute and update the models locally.

\section{Conclusion}
We propose to learn generic representation models for the UI language from instruction manuals and how-to guides. We apply prior VL models to this data and encounter new challenges due to the text richness and context sensitivity of UI screens. To address them, we build a new dataset based on the concept of functional caption, design a new vision encoder which leverages text detected in UIs, and create a new POS pre-training task. Tested on four downstream tasks, \sys largely outperforms our baselines, including a multi-task \vilbert model trained on much larger datasets.

\section*{Acknowledgements}
We thank Sahisnu Mazumder for early discussion on the \sys project and for implementing the first prototype of the \dataset web crawler. We thank Weiwei Yang for her feedback on the \sys architecture, and Kate Lytvynets for helping implement the \sys UI element detector.

\bibliography{custom,citation}
\bibliographystyle{acl_natbib}

\appendix

\section*{Ethical considerations}

One possible application of our technology are screen readers for visually-impaired users. As accessibility labels are often missing or incomplete, \sys could give them access to a much wider range of applications. In this regard, it is important that the pre-training dataset guarantees an unbiased coverage of applications and platforms. Another potential use case of \sys is task automation, which has societal and security implications: What if an agent clicks the wrong button? Is there an ``undo'' option or other recourse? A significant frontier for intelligent task completion systems is architecting them such that they can increase human productivity, yet remain amenable to human review. 

To test one of our downstream tasks, we asked four annotators to label instruction manuals of Microsoft Office applications. The annotations provided (i.e., spans of verb, object, and input entities) contain little subjectivity and ambiguity. More details on how this dataset was curated can be found in Appendix~\ref{sec:appendix-annotation}.

We have released the \dataset dataset. In line with existing pre-training data sharing policy~\cite{sharma2018conceptual, li-acl20}, we have released the collection of URLs from which we crawled the image-text pairs and the associated scripts to generate training samples. 

\section{UI data collection}
\label{sec:appendix-dataset}

As described in~\S\ref{sec:dataset}, to curate the \dataset dataset, we collect UI images from three sources. In the following we provide more details on the first (instruction manuals) and third source (general technical websites). We also describe how functional captions were generated. 

\paragraph{Instruction manual crawlers.} We built crawlers that use rules specific to a support website to extract comprehensive instruction manuals consisting of sets of instructions along with image occurrences, organized by topic and functionality. We built custom crawlers for 23 Microsoft products, which we collectively call MS apps, whose technical manuals can be found at \href{https://support.microsoft.com/<app-name>}{support.microsoft.com}, and for Zoom, whose manual can be found at \href{https://zoom.us/}{zoom.com}. 

\paragraph{Mixed data crawler.} 

We also built a crawler that uses general heuristics to extract UI images along with their alt-text and surrounding texts from a variety of support and how-to websites, for different apps and platforms (hence the name ``mixed''). The crawler first collects URLs of technical websites and how-to guides by synthesizing search queries and submitting them to web search. 

Search queries are generated in two ways. In one approach, we extract names of UI elements (e.g., ``Home button'', ``Settings icon'') from the crawled instruction manuals of MS apps (where UI elements are easily recognizable from their HTML classname attribute) and concatenate them with their corresponding app name (e.g., ``Home button, Word''). Note that to create an OOD test set, in this web crawl we avoid using names of UI elements and instructions from Teams and Zoom manuals, although overlaps due to common UI elements and instructions may occur. In a second approach, we split multi-step PixelHelp~\cite{li-acl20} instructions crawled from \href{https://support.google.com/pixelphone}{support.google.com/pixelphone} into single-step instructions. PixelHelp instructions relate to four task categories: configuring accounts, Gmail, Chrome, and Photos. 

Once search queries are generated, the mixed data crawler submits them to Google image search and parses the top 60 results. It then visits the URL of each result, and extracts images (using HTML tags) as well as preceding and subsequent text. 

We found this approach to work well for two key reasons. First, by carefully selecting the source webpages, the process ensured that the large majority of the collected images were indeed icons and UI screenshots. Second, although our seed search queries were specific to MS apps and PixelPhone, they were general enough (e.g., ``Account icon OneDrive'' derived from the OneDrive manual or  ``Tap where you can enter text'' derived from PixelHelp) to trigger results for a much larger set of apps and platforms (e.g., Apple and web).

\paragraph{Caption generation.} 
For each crawled UI image we generated from 1 up to 5 captions following simple heuristics. If the preceding text contains a set of numbered instructions we split them into individual steps, and generate captions that contain from 1 up to 5 steps. Otherwise, the entire preceding text goes into one caption. For Android support websites obtained from AndroidHowTo~\cite{li-acl20}, we used preceding or subsequent text, based on the annotations provided in the AndroidHowTo dataset. We discarded image-text pairs where the number of words in the text was less than 4 or more than 160. Overall, the average length of captions is 32 words, the average length of alt-text is 4.5 and the average length of instruction text is 29.

\paragraph{\dataset dataset summary.} 
Overall, the \dataset dataset consists of \dsize image-caption pairs. Based on the crawling source we distinguish three portions of the dataset: 1) \textbf{\crawldata} (\crawlsize pairs), containing UI data for different apps and platforms, obtained by crawling general technical websites; 2) \textbf{\androiddata} (\androidsize pairs), covering Android apps, obtained by crawling support websites listed in AndroidHowTo; and 3) \textbf{\officedata} (\officesize pairs), \textbf{\teamsdata} (\teamsize pairs), and \textbf{\zoomdata} (\zoomsize pairs), covering various MS apps and Zoom, obtained using custom instruction manual crawlers. The third category contains the highest quality data in terms of image-text alignment and instruction details. \teamsdata and \zoomdata are used only in testing, for an OOD analysis.

\section{\sys UI element detector}
\label{sec:ui-elem-detector}
The design of the \sys detector is inspired by prior work~\cite{chen2020object,remaui15,moran:redraw20} that demonstrated how off-the-shelf computer vision algorithms coupled with DNN classifiers can be successfully applied to UI screen understanding, without requiring expensive datasets of labeled UI screens. 

The \sys detector consists of two separate detection streams and a fusion layer. In one stream, it detects the edges of an image's elements using Canny's algorithm~\cite{canny86}, applied to a grey-scale map of the input image. After a dilation with a $8\times8$ rectangle kernel, it applies contour detection~\cite{SUZUKI1985-contour-map}, and then approximates contours to polygons to obtain rectangular bounding boxes. Finally, it uses non-maximum suppression (with an overlap threshold of 75\%) to eliminate overlapping boxes, and it discards boxes whose area is less than 50 pixels or whose width or height is smaller than 5 pixels. However, fully-contained boxes are not eliminated in this stage (e.g., we do not want to eliminate an icon contained in a text button). In a parallel stream, the \sys detector runs PP-OCR~\cite{du2020pp} on the input image\footnote{We discard boxes with recognized text shorter than 2 characters and with a confidence lower than 0.7.}, followed by another round of non-maximum suppression, this time only for text boxes. In the final fusion stage, the detector merges contour-identified boxes with OCR-inferred boxes. Inspired by REMAUI's approach~\cite{moran:redraw20}, we use the following heuristics: (i) we merge bounding boxes which are vertically or horizontally aligned and separated by no more than 2 pixels, (ii) we merge OCR-inferred boxes with contour-inferred boxes which fully contain them and are of similar area, and (iii) we eliminate contour-detected boxes contained in OCR-inferred boxes only if they are either smaller than 250 pixels or they are at least 85\% as big as the containing text box.

\begin{table*}[t]
\centering
\scalebox{0.85}{
\begin{tabular}{@{}llcccccc@{}}
\toprule
Task & \multicolumn{1}{c}{Dataset} &  \multicolumn{3}{c}{Split sizes} & BS & LR & NE \\
& & Train & Val & Test & & & \\ \midrule
\sys pre-training (4 tasks) & \dataset (\crawldata + \officedata + \androiddata) & \dsize & 5.6k & 7k & 128 & 4e-5 & 100 \\
UI action entailment & \dataset (\crawldata + \officedata + \androiddata) & 77k & 3.7k & 6k & 256 & 2e-5 & 50 \\
Instruction-based UI image retrieval & \dataset (\crawldata + \officedata + \androiddata) & 77k & 3.7k & 6k & 256 & 2e-5 & 50 \\
Grounding referring expressions & UIBert & 15.6k & 448 & 544 & 32 & 1e-5 & 50 \\
UI entity recognition & \androidhowto + \dataset (\officedata) & 34.5k & 4.4k & 4.5k & 32 & 1e-5 & 50 \\
\bottomrule
\end{tabular}
}
\caption{Training details including dataset sizes (reported as the number of unique image-instruction pairs), effective batch size (BS), learning rate (LR), and maximum number of epochs (NE). \dataset's \zoomdata and \teamsdata data are used only in testing, for an OOD analysis.}
\label{tb:train-details}
\end{table*}


Once bounding boxes have been identified, they are fed to a UI classifier. We fine-tune ResNeXt-101 using a collection of 263k mobile/web/Windows UI elements (obtained through data synthesis, by cropping UI elements from annotated Android UI screens~\cite{deka2017rico}, and from public icon datasets~\cite{google-material-icons}). We classify boxes into 9 classes: ``text-button'', ``image'', ``text-label'', ``image-button'', ``radio-button'', ``checkbox'', ``input-field'', ``select'', and ``other''.

The final output consists of a set of bounding boxes, each with a class label, probability vector, and recognized text (if any).

\section{Additional training details}
\label{sec:training-details}

All \sys models are implemented in PyTorch~\cite{NEURIPS2019_9015}. The text input size is 512 tokens. Alt-text is truncated to 64. The remaining tokens (at least 448) are used for the instruction text. The image region feature size is 2048. Scalar coefficients for the pre-training loss $\mathcal{L}$ are as follows: $\alpha$=0.01, $\beta$=0.8, $\gamma$=0.15, and $\delta$=0.2. 

    
In Table~\ref{tb:train-details}, we report additional details on the pre-training and fine-tuning process for each task, including datasets, training, validation and test splits, effective batch size (BS), learning rate (LR), and number of training epochs (NE). For fine-tuning tasks, NE indicates the maximum number of training epochs, as we have an early stopping using a validation split with patience of 5. Dropout is fixed at 10\% for all layers that use dropout.

\begin{figure*}[t!]
    \centering
    \includegraphics[width=1.025\linewidth]{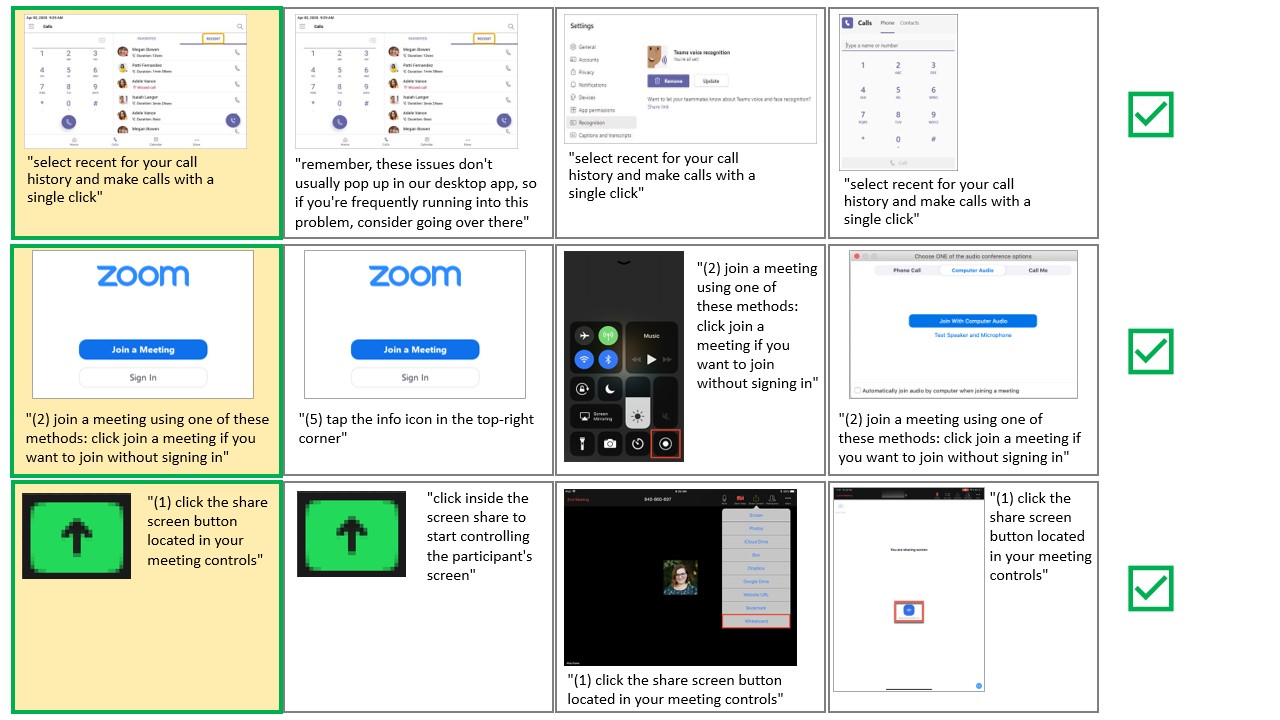}\\
    \vspace{-0.7ex}
    \includegraphics[width=1.025\linewidth]{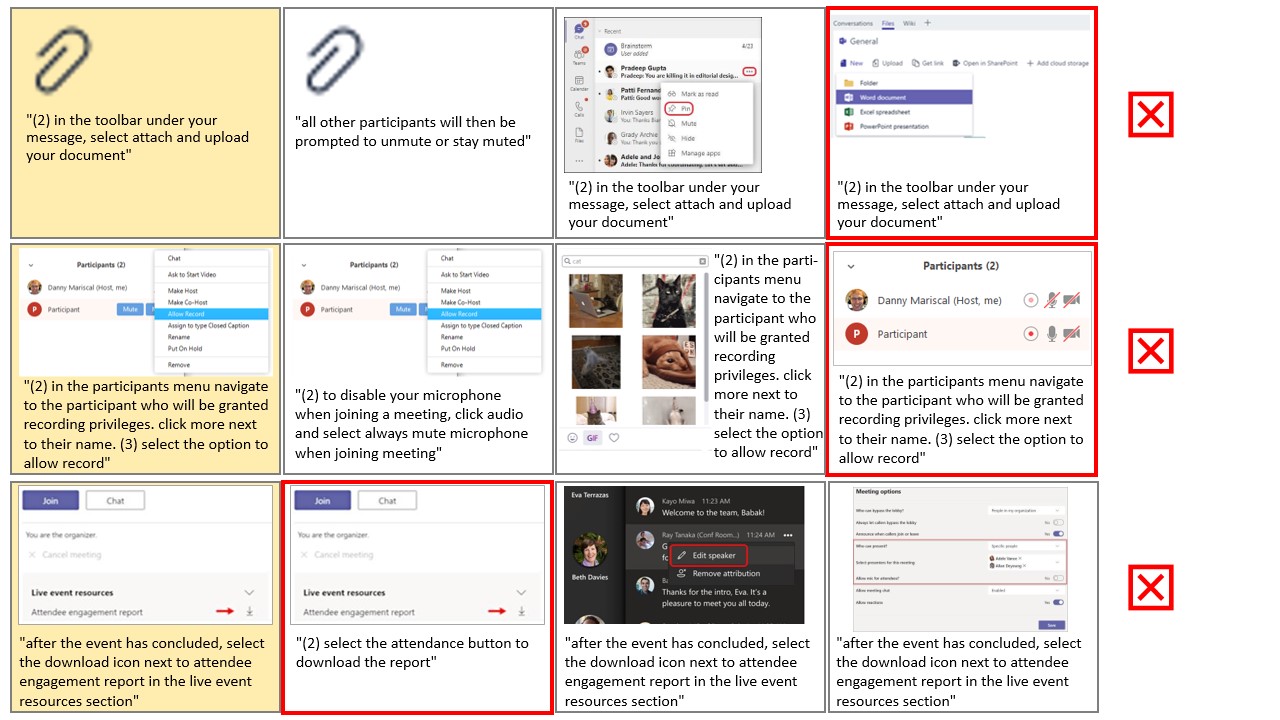}
    \caption{Samples from UI action entailment with \sys-6 predictions for Zoom and Teams. The correct pair is the first one in each row. Correct predictions are highlighted in green and wrong ones in red.}
    \label{fig:aevizerrors}
\end{figure*}

\section{Analysis of UI action entailment predictions}
\label{sec:appendix-visual-data}

In Figure~\ref{fig:aevizerrors}, we share 3 correct and 3 incorrect predictions made by \sys-6 fine-tuned for the UI action entailment task. The test dataset consists of \teamsdata and \zoomdata image-text pairs. In the figure, the true pair is always the first one, while the prediction is highlighted in green (correct) or red (wrong). The last pair is the hard negative one. Note that in the UI action entailment task, the model is not provided with the alt-text associated with the UI image, even when it is available. 

We observe that the model can select the correct image-instruction pair out of the four given choices even in cases where the alternative choices contain text or UI controls with a high overlap with the instruction text. In the first example in Figure~\ref{fig:aevizerrors} (first row), the instruction describes how to make calls in Teams; the last pair (the hard negative sample) consists of a UI screen with a call pad but the model correctly selects the first UI screen which contains the same call pad but also the ``Recent'' button mentioned in the instruction. In the second example, the instruction is about joining meetings in Zoom. While both the first and forth UI screen contain ``Join'' buttons, the model correctly selects the first UI screen which is more appropriate for the instruction given (''join a meeting'' vs. ``join with computer audio''). Finally, in the third example, the model correctly selects the first pair despite the second and forth pairs both containing a UI screen related to screen sharing and an instruction referring to the screen share function.

In cases where the model chooses incorrect pairs, we observe that it still selects very reasonable pairs. In the fifth example, \sys selects the last choice where the UI screenshot and caption are related to the participant menu; however, the correct UI screen for that caption (in the first pair) shows another view of the participant menu containing the ``more'' button mentioned in the instruction. The last example is particularly challenging: the instructions in the first and second pair, which contain the same UI screen, describe the same functionality which can be achieved through two slightly different UI controls (the ``download icon'' shown in the UI screenshot and the ``attendance button'').

\balance
\section{Labeling data for UI entity recognition}
\label{sec:appendix-annotation}
For the UI entity recognition task we collected a small dataset to train and test our models. We asked 4 annotators with technical backgrounds (2 female and 2 male, English speakers, and payed competitive hourly wages) to annotate some of the MS instruction manuals. Annotators knew their annotations would have been used for a research project. Each annotator was presented with multiple Excel files containing textual instructions for various Microsoft Office applications (e.g., Excel, Word, OneNote, Teams, etc.). Each row in each Excel sheet contained information about a specific functionality, organized as follows: (1) name of the application, (2) title of the manual section (i.e., functionality name), (3) brief description of the functionality, and (4) a set of instructions. The annotator was asked to read the set of instructions for each functionality and label the span of words that described one of three types of entity: (i) UI action verbs (by entering the \id{<verb>} and \id{</verb>} tags), (ii) UI action objects (by entering the \id{<obj>} and \id{</obj> tags}), and (iii) input parameters (by entering \id{<input>} and \id{</input>} tags). Each file was annotated by 3 annotators; in case of disagreement, a forth annotator was consulted. Before starting the annotation task, annotators were given a definition of our entity types and multiple examples, especially ambiguous ones. For instance, we showed them a sample instruction ``Specify the PowerPoint presentation that you want to open and then click...'' and labeled ``PowerPoint presentation'' as action object (rather than ``the PowerPoint presentation that you want to open'').




\section{Adapting AndroidHowTo to UI entity recognition}
\label{sec:tuple-extr}

\citet{li-acl20} propose a phrase tuple extraction (PTE) model and release the AndroidHowTo dataset, which inspired our UI entity recognition task. In PTE, given a natural language instruction the model extracts a sequence of tuples that best describe each action contained in the instruction. Each tuple consists of a UI operation (of 7 possible types, such as \id{click}, \id{input}, etc.), a UI object, and additional arguments. The goal of PTE is sequence generation. Given the instruction ``to complete your task tap A and B'', the goal of the PTE model is to output the 2 tuples \id{<operation:"click", object:"A">} and \id{<operation:"click", object:"B">}.

We build a simpler, multi-modal version of this task focused on entity extraction. We introduce equivalent entities (UI action verb, UI action object, and input parameters) and leverage \androidhowto to build a dataset to train our model. For the same example instruction above, our model's goal is entity extraction, i.e., tagging ``tap'' as an action verb, and ``A'' and ``B'' as action objects. 

An example of \androidhowto instruction with annotations is the following.

\begin{lstlisting}
    {"instructions": "Keep your phone updated: to view the most recent security update, open the Settings app and tap Security or System updates. Go to Settings > Google > Security > Google Play Protect.", 
    "operation_list": ["CLICK", "CLICK", "CLICK", "CLICK", "CLICK", "CLICK"], 
    "verb_list": ["open", "tap", "go to", "go to", "go to", "go to"], 
    "obj_list": ["the settings app", "security or system updates", "settings", "google", "security", "google play protect"], 
    "args_list": []}
\end{lstlisting}

We adapt this data to build a training dataset for our task. We keep \id{obj\_list} and \id{args\_list} intact and compute a new \id{action\_verb\_list} derived from \id{verb\_list}:

\begin{lstlisting}
    "action_verb_list": ["open", "tap", "go to"]
\end{lstlisting}

\end{document}